\documentclass[11pt]{article}

\usepackage{amsthm}

\usepackage[LGR,T1]{fontenc}
\newcommand{\textgreek}[1]{\begingroup\fontencoding{LGR}\selectfont#1\endgroup}

\usepackage{textgreek}
\usepackage[multiple]{footmisc}

\usepackage{mathrsfs}
\usepackage{amssymb}
\usepackage{euscript}
\usepackage{amsmath}
\usepackage{dsfont}
\usepackage{paralist}

\theoremstyle{definition}
\newtheorem{example}{Example}
\newtheorem{definition}{Definition}

\begin{document}

\title{On Computable Abstractions\\
\large(A Conceptual Introduction)
}
\author{Alejandro S\'anchez Guinea
\thanks{University of Luxembourg. \newline Email: ale.sanchez.guinea@gmail.com or alejandro.sanchezguinea@uni.lu}}
\date{}

\maketitle

\begin{abstract}
This paper introduces abstractions that are meaningful for computers
and that can be built and used according to computers' own
criteria, i.e., computable abstractions. It is analyzed how 
abstractions can be seen to serve as the building blocks
for the creation of one own's understanding of things, which
is essential in performing intellectual tasks.
Thus, abstractional machines are defined, which following a 
mechanical process can, based on computable 
abstractions, build and use their own 
understanding of things.
Abstractional machines are illustrated through an
example that outlines their
application to the task of natural language processing.
\end{abstract}

\section{Introduction}

An {\em abstraction} may be seen as a construct that encompasses one or many 
things, holding `meaningful information' about the things encompassed according to 
certain `criteria'. 

It is the meaningful information of an abstraction what makes possible to `discard'
or `forget about' 
the things encompassed.
Suppose, for instance, that we have a computer program named {\tt print\_on\_screen} 
for printing something on screen. 
The name of the program may tell us --- as it is the case here --- 
what the program is about (i.e., it gives 
us meaningful information about the program). 
Based on this we can use the program (e.g., to build other programs) irrespectively 
if we know or not what are the operations that the program needs to execute to print 
something on screen.
That is, the name of the program together with its meaning form a 
construct that is abstracting the program including the operations that it uses to 
work as expected.

Abstractions are essential when a person performs intellectual tasks. 
Based on their meaningful information they can serve as the building
blocks for creating one's own understanding
of things, allowing to distinguish one thing from others regardless of their level of 
abstraction, as well as to relate different
things under different contexts and at different levels of abstraction. In this way, humans
can figure out or learn what things are about (e.g., what is to be done, how to do it,
what can be used to do it) and act accordingly.

While computers are typically in close interaction with abstractions (in any computer
program), abstractions that have been defined and used in the past cannot be
considered to be computable, as their information has remained meaningful only 
for humans and the criteria to build them have remained dependent on human thinking.
These issues have limited the possibilities of computers towards performing 
intellectual tasks by not having the capability of creating their own understanding of things. 
One can observe that regardless of how
`data-dependent' or `data-driven' computers have been made, they have remained being
governed by the original program(s), perhaps with modified parameters, but never
able to make sense out of things.

This paper introduces computable abstractions which are meaningful for computers
and that can be built and used according to computers' own criteria.
Based on this, it is proposed the construction of abstractional machines capable of 
building and using their own understanding of things. 
{\S\,\ref{sect:analysis}} presents a preliminary analysis that will
serve for the development of further sections, in particular, {\S\,\ref{sect:absandcomp}}
analyzes  current approaches, their issues, and outlines the solution proposed in this paper.
{\S\,\ref{sect:compabs}} introduces computable abstractions and describes
a mechanical abstracting process to build and use computable abstractions.
In {\S\,\ref{sect:abs_machines}} abstractional machines are defined and
illustrated through an example that outlines their application to the task of natural
language processing.

\section{Preliminary Analysis} \label{sect:analysis}

\subsection{Abstractions \& Computers} \label{sect:absandcomp}

In this section we analyze the view about abstractions
and the mechanisms to obtain them and manipulate them, which has been widely
accepted in artificial intelligence approaches and in computer science, in general.
We explain the issues that such view entails and outline the solution that is
proposed in this paper. 

\subsubsection{Current View}

\paragraph*{Abstraction: The Object}

Abstractions that interact with computers have typically been designed either
as to ease human interpretation or based on what humans understand, and 
not as to being meaningful for computers (e.g., 
object-oriented programming \cite{booch1986object}, 
ontologies \cite{chandrasekaran1999ontologies}, 
logic \cite{nilsson1991logic, mccarthy1987generality}).

\begin{example} \label{ex:oo1}
Consider an object-oriented class {\tt C}, in some computer $\mathds{C}$, 
which was designed by some person \textgreek{r} as part of the writing of program {\tt P} 
in $\mathds{C}$. We assume that {\tt C} was designed with the intention of representing 
some category of things \textgreek{k} in the context of {\tt P}, as understood by \textgreek{r} 
(at least).
Thus, the specification of {\tt C} is in general expected to help us (humans), specifically 
those who understand it, to reason about \textgreek{k} in the context of {\tt P}, to ease 
the overall understanding for \textgreek{r}, etc. 
Nonetheless, for $\mathds{C}$ such specification provides only information related to, 
for instance, how memory is to be reserved for the creation of instances of {\tt C}, 
and not about the category that {\tt C} is supposed to represent (i.e., \textgreek{k}).
\end{example}
The term `computable abstraction' has appeared scarcely in the literature, e.g., 
\cite{mostow1989discovering} in artificial intelligence, and 
\cite{mine2006symbolic, cousot2013theories, johnson2014abstracting} in formal verification and 
model checking. Nonetheless, different works
have considered (implicitly) an abstraction to be computable inasmuch as it can be processed
by a computer, e.g., \cite{giunchiglia1992theory, knoblock1994automatically,
chittaro2004hierarchical, backstrom2012abstracting}. 

\paragraph*{The Abstraction Mechanism}

Approaches in artificial intelligence have employed a so called 
abstraction mechanism \cite{saitta2013abstraction}.
The prevailing view of an abstraction mechanism
(e.g., in knowledge representation \cite{hobbs1985granularity}, 
reinforcement learning \cite{barto2003recent}, 
heuristic methods \cite{bulitko2007graph}), can be 
pictured as a program {\tt P}
that takes some input {\tt I} and obtain, according to a predefined `criteria', some 
`representation' which relates objects in {\tt I} between each other or relate them
to other objects given by the program, where the resulting `representation' is supposed to
be easier to use or somehow more convenient than the original input. A simple example of this 
can be a program for classification 
that has been devised to classify a set of objects
into some predefined classes.

\subsubsection{Issues in Current View}

The following issues can be observed in the `current view' with regard to the goal of
having machines capable of performing intellectual tasks with
results that can be comparable to humans' results.

\begin{description}
  \item[Issue 1.] Abstractions are not designed as to be meaningful for computers
  \item[Issue 2.] The `criteria' employed by an abstraction mechanism depend on humans to be devised
  and adapted to each problem or class of problems and specific situation
  \item[Issue 3.] The resulting construct(s) (or `representation') of
  an abstraction mechanism can only be understood (or need to be interpreted) by humans.
  \item[Issue 4.] {Issue 1}, {Issue 2}, and {Issue 3} imply that currently the abstractions that 
  interact with computers are not computable (at least not as a whole), and there is no
  defined mechanism that allows computers to build and use abstractions based on their 
  own criteria.
  This, in turn, prevent the development of 
  mechanisms that allow computers to build and use
  their own `understanding' of things, limiting their possibilities towards performing intellectual tasks
  on their own, by being inevitably dependent on humans' criteria and understanding.
  \item[*] `Computer understanding' mentioned in {Issue 4} has also a `current view'
  (as part of artificial intelligence development), which is focused mainly in 
  information-seeking tasks such as text classification, information retrieval, and information
  extraction \cite{russell2009artificial}. Approaches in this area have attempted to
  simulate inasmuch as possible an understanding of natural language through the
  use of, for instance, probabilistic and statistical methods 
  (e.g., \cite{joachims2001statistical, collins2005discriminative, collobert2011natural}).
  However, the models used in such approaches are not computable abstractions, 
  and thus the issues above hold for them as well.
\end{description}

\subsubsection{Proposed Solution}

To deal with the issues listed above we first, in {\S\,\ref{sect:opmachines}},
analyze the capabilities of computers and the kind of objects that they
have available. Based on this analysis, we introduce in {\S\,\ref{sect:compabs}}
abstractions that can be built from the objects available for computers, based on the 
computers' capabilities, and following criteria general enough as to
be applied by any computer, to any problem, in any situation. 
{\S\,\ref{sect:abs_machines}} defines, with regard to our notion
of understanding provided in {\S\,\ref{sect:understanding}},
abstractional machines capable of building and using their
own understanding of things (built as a system of computable abstractions)
to perform their tasks. 

\subsection{Operating Machines} \label{sect:opmachines}

Here, we analyze the kind of objects available for computers,
in order to later, in {\S\,\ref{sect:compabs}}, be able to define what kind of meaningful
information can be obtained from such objects, by computers.
Some considerations related to how computers interact with the outside
world are also given, as they will be necessary later in the paper.

Let us call {\em operating machine} to any model operationally equivalent 
to a deterministic Turing machine. Examples of operating machines
are, among others, Knuth's model \cite{knuth1999mmixware}, 
Kolmogorov machines \cite{uspenski:algorithms}, programming languages 
(e.g., C, C++, Python), and operating systems.

Let us call 
{\em machine operation} or simply {\em operation} to any action that
has been specified in an operating machine $\mathds{M}$ and
can be executed by $\mathds{M}$, i.e., it is a special kind of action
that has been precisely and unambiguously specified for an operating 
machine. We say that an operation is {\em in} some specific operating
machine if it is executable in that machine, that is, the operation
has been specified in that machine.

One can observe that all constructs that exist in or are used by an
operating machine are operations, which the machine sees only as
actions that are to be performed. Such constructs include procedures
(sometimes known as routines, methods, etc.), data structures, data
abstractions \cite{cardelli1985understanding}, user-defined types, and, 
in general, any construct that
has been defined in an operating machine. Even constructs that 
are defined as to carry meaning such as 
knowledge representations \cite{davis1993knowledge}
are for the machine only operations. We can see that, currently, 
regardless of the meanings that they may have for 
us (humans), those constructs only tell to the machine what to do 
and how to do it, and the machine ought to follow such directions.

The specification of an operation in some operating machine may be 
seen as to be given by a unique identifier \footnote{This identifier should not be
understood as a regular program's name but rather as 
what {A.M. Turing} defined as {\em description number} in \cite{turing1936computable}} 
(unique for $\mathds{M}$) 
and a set or a sequence of operations in $\mathds{M}$, which will be
referred herein as the composition of the operation. 
This unless the operation lies in the lowest operational level of $\mathds{M}$,
in which case only the unique identifier is considered, and the set or sequence
of operations is assumed to exist, however, out of the scope of $\mathds{M}$ 
 (e.g., operations `{{\tt scan \!a \!symbol\! from\! the\! tape}}' and 
 `{{\tt print \!a \!symbol \!on \!the \!tape}}' are in the lowest 
 operational level of a Turing machine \cite{turing1936computable}). 
 Herein, we denote an operation $\varphi_{a}$ specified by a set of operations as 
 ${\varphi_{a}\colon\{ \varphi_{1}, \dots, \varphi_{n} \}}$
 and an operation $\varphi_{b}$ specified by a sequence of operations as
 ${\varphi_{b}\colon\langle \varphi_{1}, \dots, \varphi_{n} \rangle}$.
 
In what respects to possible interactions between an operating machine 
and its outside world, it can be said that an operating machine can recognize 
(or process) an object outside if there is an executable operation in the machine 
that is uniquely associated with the object outside.\footnote{The consideration is 
neither about the operation(s) that may work (or interact) with the object nor about 
the operation(s) in charge of process it; it is about the object in the machine that refers 
to the object outside.}
As examples, one may consider any input given to an operating machine; 
it is not hard to notice that such input needs a corresponding executable operation 
in the machine in order to be recognized (or processed) by the machine.

\subsection{Understanding} \label{sect:understanding}

In this section, we describe aspects that can be observed on humans' understanding, 
which are considered in {\S\,\ref{sect:abs_machines}} for shaping 
the understanding of the proposed abstractional machine. 
\begin{enumerate}[I.]
\item {\em Understanding} an object may be regarded as to have
meaningful information about that object for some `specific purpose'. 
\item We say that an understanding of an object \textgreek{e} combines together
abstractions related to \textgreek{e} (each with meaningful information) 
in a system of abstractions, 
which provides the capability of distinguishing
\textgreek{e} from other objects, as well as to relate it to different
objects, all this, potentially, in different contexts (situations, or for different purposes)
and involving objects at different levels of abstraction. The understanding or system of abstractions
is adapted and grown as more meaningful information is gathered
\item Meaningful information may appear at different abstraction levels, 
where high-level information is built from primitive information. 
Meaningful information is considered primitive if either it is accepted without need for further 
understanding (i.e., as axioms) or it can be drawn by executing known actions. 
One may consider, for instance, a primitive action such as `{\slshape sliding 
a pencil over a piece of paper}', which, even without understanding any of the high-level ideas 
involved (e.g., what a pencil is), it allows to describe information about the pencil and the 
paper. 
That is, if we use \textgreek{a} as the identifier of the action `{\slshape to 
slide over}', $\text{\textgreek{e}}_{1}$ for `{\slshape a pencil}', and 
$\text{\textgreek{e}}_{2}$ as `{\slshape a piece of paper}', executing 
\textgreek{a} using  $\text{\textgreek{e}}_{1}$ on  $\text{\textgreek{e}}_{2}$ can provide 
information about the objects involved, e.g., \textgreek{a} can use 
$\text{\textgreek{e}}_{1}$ and act on $\text{\textgreek{e}}_{2}$, and
$\text{\textgreek{e}}_{1}$ can interact with $\text{\textgreek{e}}_{2}$ and is operable 
within \textgreek{a}. 
Furthermore, considering another simple action such as `{\slshape to 
notice changes in the color of} $\text{\textgreek{e}}_{2}$', it would be possible to draw 
information that could begin building an understanding on the high-level idea of 
`{\slshape painting}'.
\item Understanding is created as it is needed 
and the amount of information in someone's understanding is based on what that 
someone has to do. 
For instance, one may consider that a person has sufficient understanding about some 
topic when that person is able to solve exercises or tests on such topic. 
\item Each individual builds its own understanding. In order for 
two individuals to understand each other, i.e., achieve
a {\em common understanding}, the two respective 
understandings need not to be the same but simply to share their purpose (e.g., the purpose
 of a person when saying ``hello'' is to greet someone).
\end{enumerate}

\section{Computable Abstractions} \label{sect:compabs}

From the analysis about operating machines in {\S\,\ref{sect:opmachines}}, 
it should be clear that  
to be suitable for machines, abstractions have to be built or derived from the 
specifications of operations following a process that involves nothing but the 
execution of operations (i.e., a mechanical process). To this end, we find a criteria
for building abstractions that any machine can apply to any problem, in any situation.

\subsection{Computable Concepts}


Based on the specifications of operations, it is possible to describe 
``operabilities'' that an operation satisfies with respect to other 
operations. 
An ``operability'' should be understood as the ``ability'' to be 
operable in a particular way.

``Operabilities'' are indeed abstractions, since an ``operability'' 
may encompass one or many objects (those of which are operable 
in the way defined by the ``operability'') and holds meaningful information 
about the objects it encompasses, namely, one way in which they can be 
operable. 
Nonetheless, not all ``operabilities'' that can be described from the 
specifications of operations can be processed by a machine 
(i.e., not all are computable).
In fact, many ``operabilities'' that could be intuitively recognized are 
not computable as they depend on human thinking to be understood.
For instance, based on an operation {\tt print}, we (humans) could 
intuitively recognize the ``operability'' `{\slshape printable}'. 
However, in order to recognize such ``operability'' it is necessary first 
to have an understanding of what it means to print something 
and then to understand what it means to be `{\slshape printable}'. 
As it is, an operating machine cannot do this, since for the machine 
the operation {\tt print} is (as any other operation) just an action that 
is to be executed following its specification (i.e., the machine
does not hold any meaning of its operations).

In order to describe computable ``operabilities'' from the specifications 
of operations, it is necessary to consider only the aspects 
that an operating machine can process from such specifications. 
These are: 
\begin{enumerate}[\itshape i\upshape)]
\item an operating machine can distinguish between its operations 
based on their unique identifiers; 
\item an operating machine has the description of the composition 
of its operations (those that do not lie on its most elementary 
operational level), including the identifiers of the operations involved 
and, for the case of a sequence, their order of execution
\end{enumerate}.

\begin{definition}
For an object $x$ in some operation $\varphi$ in an operating 
machine $\mathds{M}$,
we call {\em computable concepts} to the ``operabilities'' that 
$\mathds{M}$ can
define about $x$ with respect to $\varphi$ and its components.
Computable concepts, which we will denote by $\EuScript{C}_{i}$ 
($i$ being the unique identifier of the concept for $\mathds{M}$),
can be of two types:
\begin{enumerate}[{[}Type I{]}]
  \item \label{it:first} such that refers to 
  `{\slshape being operable within the specification of some 
  particular operation $\varphi$}',
  denoted as $\EuScript{C}_{i}(x)\colon\varphi(x)$ or simply 
  as $\EuScript{C}_{i}\!\colon\varphi(x)$
  \item \label{it:second} such that refers to 
  `{\slshape being operable within some particular set of 
  operations}' or, 
  for the case of sequences, `{\slshape ... in a specific position 
  of a sequence of operations}';
  for instance, for a composition sequence $\langle \varphi_{1}, 
  \varphi_{2}\rangle$ of
  some operation, a concept $\EuScript{C}_{j}$ that refers 
  to `being operable in the
  first position of a two-element sequence, where the second 
  element is $\varphi_{2}$'
  can be described by the machine, being denoted as 
  ${\EuScript{C}_{j}(x)\!\colon \langle x, \varphi_{2}\rangle}$.
\end{enumerate}
\end{definition}

A concept may encompass more than one object, 
all of which satisfy the operability that the concept represents. 
This defines an equivalence relation between the objects that satisfy the concept. 
Thus, for instance, if we have a concept defined as 
${\EuScript{C}_{a}\!\colon \varphi_{a}(x)}$ and
two objects $a$ and $b$ that satisfy this concept, which we denote
as ${a\!\dashv\!\EuScript{C}_{a}}$ and ${b\!\dashv\!\EuScript{C}_{a}}$, then
it follows that $a$ and $b$ are equivalent with respect to
$\EuScript{C}_{a}$, i.e., ${a \overset{\EuScript{C}_a}{\sim} b}$.

Concepts may entail, as well, relations that are not necessarily 
equivalence relations. 
For instance, if we know that $\varphi_{a}$ is operable within the 
specification of an operation $\varphi_{b}$ and we define 
${\EuScript{C}_{e}\!\colon \varphi_{b}(x)}$, we can say that 
${\varphi_{a}\!\dashv\!\EuScript{C}_{e}}$. 
And, since we have ${a\!\dashv\EuScript{C}_a}$, 
with ${\EuScript{C}_{a}\!\colon\varphi_{a}(x)}$, 
we have then a relation between the object $a$ and 
$\EuScript{C}_e$, i.e., 
${a\!\overset{\EuScript{C}_a}{\longrightarrow}\!\EuScript{C}_e}$, 
with which $a$ is related to any object that satisfies $\EuScript{C}_e$.

An object in an operating machine may satisfy in general more than 
one concept, through which it may get related to other operations and concepts. 
When the object is indeed operating within the specification of a particular 
operation, one specific concept gets satisfied and the extent of relations covered 
in general by the object gets {\em restricted} to a subset of operations and concepts 
that are related to the concept that is being satisfied. 
For instance, for $a$ above, if we have that, in addition to $\EuScript{C}_a$, 
$a$ satisfies $\EuScript{C}_{f}$ (i.e., ${a\!\dashv\!\EuScript{C}_{f}}$) 
through which it is related to some objects $c$ and $d$. 
Thus, we have 
${a\!\overset{\EuScript{C}_a}
{\longrightarrow}\!\underline{b}, \EuScript{C}_e}$,\footnote{Equivalent 
objects appear underlined.} and ${a\!\overset{\EuScript{C}_f}{\longrightarrow}\!c, d}$, 
and in general ${a\!\longrightarrow\!\underline{b}, \EuScript{C}_{e}, c, d}$. 
Then, when in an execution $a$ is operating in such a way as to satisfy 
$\EuScript{C}_a$ only (i.e., operating within $\varphi_{a}$), the extent of relations 
of $a$ gets restricted accordingly, that is, 
${a\!\!\restriction\!\EuScript{C}_{a}\!\longrightarrow\!\underline{b}, \EuScript{C}_{e}}$.

Computable concepts are valid for any executable object in an operating machine, 
since all such objects can be operable in at least one way given that they all 
can be executed and, thus, can be part of the composition of some operation. 
Furthermore they allow to: 
\begin{enumerate}[\itshape i\upshape)]
  \item account for levels of abstractions based on operational levels 
  ({Type~\ref{it:first}});
  \item deal with contextual information ({Type~\ref{it:first}} and 
  {Type~\ref{it:second}});
  \item differentiate objects, since, in general, the only object that can 
  be operable in all the same ways as some object $x$, is the same 
  object $x$;
  \item relate objects based on computable concepts of {Type~\ref{it:first}}, 
  that is, all operations that are part of the composition of an operation 
  can be said to be related;
  \item relate objects based on computable concepts of {Type~\ref{it:second}}, 
  that is, all operations that appear surrounded by the same operations 
  within two different compositions can be said to be related. 
\end{enumerate} 

\subsection{Mechanical Abstracting Process}

Given some executable operation $\varphi$ in an operating machine
$\mathds{M}$, a computable abstraction of $\varphi$ is obtained by describing
$\varphi$ based on the computable concepts that $\varphi$
can be found to satisfy in `selected' 
operations in $\mathds{M}$. The criteria to decide on what operations
to select for the derivation of computable concepts depends on what we called 
`specific purpose' in the notion provided about `understanding' in 
{\S\,\ref{sect:understanding}} and it is not part of the abstracting process.
(Abstractional machines presented in {\S\,\ref{sect:abs_machines}} 
entail this `specific purpose' in their sets of tasks).

For instance, say we want to abstract (through a mechanical process)
an operation $\varphi_{1}$ that is in an operating machine $\mathds{M}$,
based on operations $\varphi_{a}\colon\langle \varphi_{1}, \varphi_{2}\rangle$
and $\varphi_{b}\colon\langle \varphi_{3}, \varphi_{1}, \varphi_{4}\rangle$, with
$\varphi_{a}, \varphi_{b} \in \mathds{M}$. Then, we describe the concepts
that $\varphi_{1}$ satisfies in $\varphi_{a}$ and $\varphi_{b}$. These are
$\EuScript{C}_{1}\colon\varphi_{a}(x)$, 
$\EuScript{C}_{2}(x)\colon\langle x, \varphi_{2}\rangle$,
$\EuScript{C}_{3}\colon\varphi_{b}(x)$, and
$\EuScript{C}_{4}(x)\colon\langle \varphi_{3}, x, \varphi_{4}\rangle$.
Finally, the abstraction of $\varphi_{1}$ stays as
\begin{equation} \label{eq:abstraction}
  \varphi_{1}\dashv \EuScript{C}_{1}, \EuScript{C}_{2}, \EuScript{C}_{3}, \EuScript{C}_{4}
\end{equation}
About this abstraction we should note:
\begin{itemize}
  \item \eqref{eq:abstraction} is an abstraction entirely meaningful for $\mathds{M}$,
  including the object being abstracted (i.e., $\varphi_{1}$), and the meaningful information 
  about it, namely, the concepts that $\varphi_{1}$ satisfies (which are meaningful for $\mathds{M}$).
  That is, \eqref{eq:abstraction} is a computable abstraction. We can see that the machine itself 
  has the explanation of the abstractions involved (e.g., `{\slshape $\varphi_{1}$
  is encompassed by $\EuScript{C}_{1}$ because it is operable in $\varphi_{a}$}'). 
  Furthermore, the process do not require human thinking to create more
  abstractions or to explain them, it only requires to acquire (or abstract)
  more information.
  \item Based on \eqref{eq:abstraction} $\varphi_{1}$ can be used by the machine 
regardless if its specification is available or not, that is, the computable abstraction
of $\varphi_{1}$ allows the machine to, for instance, `forget about' the specification of $\varphi_{1}$
  \item $\varphi_{1}$ as well as any element of its abstract description 
and its description as a whole are seen as separate objects. 
That is to say, the process of mechanically abstracting $\varphi_{1}$ does not imply to modify 
$\varphi_{1}$, neither in general nor for the machine for which is an abstraction. 
Instead, the process yields a computable description of $\varphi_{1}$, which can potentially be 
associated with other objects (provided that they satisfy such description) 
and serve to build other descriptions. 
\end{itemize} 

To describe a computable abstraction of an object outside some machine 
$\mathds{M}$ it is necessary to use objects in $\mathds{M}$ that are associated 
with objects outside. 
Let us first recall from {\S\,\ref{sect:opmachines}} that any object outside $\mathds{M}$ 
can be said to be recognizable 
(or processable) by $\mathds{M}$ if there is an executable operation in $\mathds{M}$ 
that is uniquely associated with that object. 
Thus, in order to describe a computable abstraction, the concrete object 
outside $\mathds{M}$, let us call it \textgreek{e}, has to be uniquely associated with an 
executable operation in $\mathds{M}$.
Then, \textgreek{e} is abstracted by considering the computable concepts that its associated object, in 
$\mathds{M}$, satisfies with respect to objects also in $\mathds{M}$, which are themselves 
associated with the objects outside $\mathds{M}$ that constitute the context of interest for 
abstracting \textgreek{e}.

\section{Abstractional Machines} \label{sect:abs_machines}

\begin{definition}
An {\em abstractional machine} is a machine that follows a mechanical 
abstractional process through which it builds and uses a system of 
computable abstractions (which constitutes its understanding of things)
in order to perform its tasks.
\end{definition}

For an abstractional machine $\widetilde{\EuScript{M}}$ with a system 
of abstractions $\widetilde{\EuScript{K}}$, an operating machine 
$\mathds{M}$ serves as the foundation from which $\widetilde{\EuScript{M}}$
is defined.  In addition, $\mathds{M}$ is on charge of executing all operations that 
$\widetilde{\EuScript{M}}$
needs to perform as part of its mechanical abstractional process.
Thus, $\widetilde{\EuScript{M}}$ should not be seen as a model of computation but
rather as a model that, based on an existing operating machine, performs a particular kind
of mechanical process, that is, one that manipulates abstractions building its
own understanding of things based on them.
 
The definition of $\widetilde{\EuScript{M}}$ comprises the definition of the set $\Pi$ of tasks 
(which are executable operations in $\mathds{M}$) that are to be performed by 
$\widetilde{\EuScript{M}}$ and the specification of how $\widetilde{\EuScript{K}}$ 
is to be built, including what is considered to be the ground of $\widetilde{\EuScript{K}}$ 
and what are the abstraction levels of interest for $\widetilde{\EuScript{M}}$.
The set $\Pi$ of tasks can be seen as to entail the `specific purpose' for
which the understanding of $\widetilde{\EuScript{M}}$ is being built.

\subsubsection*{System of Computable Abstractions}

The {\em system of computable abstractions} $\widetilde{\EuScript{K}}$ 
of $\widetilde{\EuScript{M}}$
is built as an abstractional model, with its lowest abstraction 
level and criteria of decomposition established as part of the definition 
of $\widetilde{\EuScript{M}}$.

We call {\em abstractional model} to a construction of interest composed of
a collection of abstractions that comply with a given criteria of validity, where
there cannot be contradiction between any two abstractions in the model
(as for the the criteria of validity). The construction of an abstractional model
requires to establish an initial collection of abstractions that are considered
of foundational importance and are assumed to be non-contradictory between 
each other. We call this the {\em ground} of the model. In addition, the abstraction
levels that are of interest for the model have to be established. First, the lowest 
abstraction level (or elementary level) establishes the point from which no further 
decomposition of abstractions is considered within the model. 
Then, a criteria of decomposition establishes the components that will be of interest 
for the model when considering each object. In order to build new abstractions 
in the model, only abstractions from its ground or that have been derived from its
ground can be used. Therefore, a criteria of validity imposed over abstractions in 
the ground of
a model will extend to all its derived abstractions, thus making the whole model
consistent under such criteria.

The reason to build $\widetilde{\EuScript{K}}$ as an abstractional model is to ensure
its the consistency and the regular behavior of $\widetilde{\EuScript{M}}$.
Thus, the behavior of an abstractional machine depends on the current state of its system of 
abstractions.
If we consider an abstractional machine performing tasks in which its system of 
abstractions do not suffer any modification, we will observe regular behavior. However, 
if due to the abstractional process its system of abstractions gets modified, then the behavior 
of the machine might vary accordingly. 
In spite of this, since the system of abstractions is built as an abstractional model, it is 
ensured that the construction will happen in a consistent manner, i.e., abstractions in 
different stages of the construction of the system of abstractions will never contradict each other.

\subsubsection*{Mechanical Abstractional Process}

To describe the mechanical abstractional process followed by an abstractional machine
$\widetilde{\EuScript{M}}$ we can picture it as a program $\widetilde{{\tt P}}$ of an 
operating machine 
$\mathds{M}$. Thus, we say that $\mathds{M}$ is expected to deal with
a set of tasks $\Pi$ in which creating and using an understanding of the input
is important or can be regarded as beneficial. To deal with $\Pi$
we define a program $\widetilde{{\tt P}}$ in $\mathds{M}$ of
a particular kind: As any program, $\widetilde{{\tt P}}$ is specified to 
perform some particular tasks,
in this case $\Pi$. However, (differently from other type of programs)
$\widetilde{{\tt P}}$ will build and use its own understanding of the input
that needs to be understood to perform $\Pi$. Roughly, this process is 
performed by $\widetilde{{\tt P}}$ by attempting to combine in a consistent 
construction its initial understanding of things (e.g., problem definition,
base rules, or previous knowledge on similar kind information) with the 
understanding that can obtain by abstracting, through computable
abstractions, the objects from the input involved in $\Pi$.

The process of including abstractions to the system of abstractions $\widetilde{\EuScript{K}}$ 
of $\widetilde{\EuScript{M}}$
involves checking such abstractions against related objects in $\widetilde{\EuScript{K}}$.
This includes to check the relations that follow from the abstractions which, if
$\widetilde{\EuScript{M}}$'s definition states it, should be included into $\widetilde{\EuScript{K}}$.
The abstractions and consequent relations will be included into $\widetilde{\EuScript{K}}$
only if their inclusion preserves the consistency of $\widetilde{\EuScript{K}}$ (as of
the abstractional model construction).

\subsection{Example}

We consider the task of processing natural language. 
We observe that most, if not all, activities within this 
task (e.g., to provide a definition of some word, 
to provide a summary or a translation of some text) 
can be reduced to what we shall call a ``contextual relation''; 
that is, given a linguistic construction \textgreek{l}, 
find a linguistic construction \textgreek{l}' that is related to 
\textgreek{l}, within some context \textsigma~(i.e., 
${\text{\textgreek{l}}\!\overset{\text{\textsigma}}
{\longrightarrow}\!\text{\textgreek{l}'}}$).

First let us consider a very basic way in which an 
abstractional machine could reach an understanding of some natural 
language. Say we have an abstractional machine
$\widetilde{\EuScript{M}}$ that has a task $\pi_{1}$
aimed to read texts that are assumed to be meaningful
in some natural language
 $\mathscr{L}$ and abstract them putting them into its 
 system of abstractions $\widetilde{\EuScript{K}}$,
 which in this case has on its ground only the
 assumption that all texts that being read are meaningful.
 Based on this simple task $\widetilde{\EuScript{M}}$
 can build and understanding of linguistic constructions
 in $\mathscr{L}$, which allow $\widetilde{\EuScript{M}}$
 to relate one linguistic construction to others, as well
 as to differentiate it from other linguistic constructions. Thus, 
 the  understanding of $\widetilde{\EuScript{M}}$ will not
 depend on human thinking but only on the information
 that it can gather, i.e., the more texts $\widetilde{\EuScript{M}}$
 reads the broader its understanding can be.
 Let us consider a more concrete case. Suppose 
 the sentence ``He is a good man'' is given to $\widetilde{\EuScript{M}}$, 
and it is processed by the operating machine $\mathds{M}$ 
that serves as $\widetilde{\EuScript{M}}$'s foundation as 
${\varphi_{s}\colon\langle\varphi_{1}, \varphi_{2}, 
\varphi_{3}\rangle}$, with $\varphi_{1}$ associated with 
``He is a'', $\varphi_{2}$ with ``good'', and $\varphi_{3}$ 
with ``man''. 
From this, $\widetilde{\EuScript{M}}$ abstracts 
$\varphi_{2}$ as ${\varphi_{2}\!\dashv\!\EuScript{C}_{1},\! 
\EuScript{C}_{2}}$, where ${\EuScript{C}_{1}\colon\varphi_{s}(x)}$ 
and ${\EuScript{C}_{2}(x)\colon\langle\varphi_{1}, x, \varphi_{3}\rangle}$. 
Furthermore, let us consider that a sentence ``He is a bad man'' 
is also given, and $\mathds{M}$ process it as 
${\varphi_{s}'\colon\langle\varphi_{1}, \varphi_{2}', \varphi_{3}\rangle}$, 
with $\varphi_{2}'$ being associated with ``bad''.
From here, $\widetilde{\EuScript{M}}$ abstracts 
$\varphi_{2}'$ as ${\varphi_{2}'\!\dashv\!\EuScript{C}_{3},\!\EuScript{C}_{2}}$, 
where ${\EuScript{C}_{3}\colon\varphi_{s}(x)}$.
Based on the computable abstractions of 
$\varphi_{2}$ and $\varphi_{2}'$, $\widetilde{\EuScript{M}}$ 
can describe the relation ${\varphi_{2}\!\overset{\EuScript{C}_{2}}
{\longrightarrow}\!\varphi_{2}'}$ (in fact, 
${\varphi_{2} \overset{\EuScript{C}_{2}}{\sim} \varphi_{2}'}$), 
which is indeed a ``contextual relation'' at a primitive abstraction level. 

The more linguistic constructions $\widetilde{\EuScript{M}}$ finds 
in which $\varphi_2$ is operable, the more complete and narrower 
the description of $\varphi_{2}$ will be for $\widetilde{\EuScript{M}}$, 
based on the concepts that $\varphi_{2}$ satisfies and the objects 
(concepts and other operations) that are related to it through those concepts. 
For instance, let us now suppose that the sentence 
``He is very kind and generous'' is given and identified as $\varphi_{t}$. 
In addition, let us eliminate from consideration the possibility of 
the idea of {\slshape `irony'}, which would allow $\varphi_{t}$ 
to appear next to $\varphi_{s}'$ meaningfully.
Thus, $\widetilde{\EuScript{M}}$ can find 
${\varphi_{u}\colon\langle\varphi_{s}, \varphi_{t}\rangle}$ 
(but not a linguistic construction with composition 
${\langle\varphi_{s}', \varphi_{t}\rangle}$). 
From here, $\widetilde{\EuScript{M}}$ abstracts 
$\varphi_{s}$ as ${\varphi_{s}\!\dashv\!\EuScript{C}_{\beta}}$, 
where ${\EuScript{C}_{\beta}(x)\colon\langle x, \varphi_{t}\rangle}$~\footnote{
The other type of concept is omitted since it is not relevant for this example.}, 
and then it abstracts $\varphi_{2}$ accordingly, i.e., 
${\varphi_{2}\!\overset{\EuScript{C}_{2}}{\longrightarrow}\!\EuScript{C}_{\beta}}$.
Therefore, $\varphi_{2}$ is singled out from $\varphi_{2}'$, since 
$\varphi_{s}'$ cannot not be found satisfying $\EuScript{C}_{\beta}$ in this case.

The understanding that $\widetilde{\EuScript{M}}$
obtains of $\mathscr{L}$ based on $\pi_{1}$, another task $\pi_{2}$ can
be defined for $\widetilde{\EuScript{M}}$ to check
if some texts given as inputs are meaningful or not (according to
$\widetilde{\EuScript{M}}$' understanding). This could be made more
complete by including in the ground of the system of abstractions of the machine
computable abstractions of grammar rules of $\mathscr{L}$.

The kind of relation we skipped by avoiding the idea of {\slshape `irony'} 
above, i.e., a ``contextual relation'' through a high-level context (or idea) 
(e.g., ${\varphi_{s}\!\overset{\text{{\tiny\slshape irony}}}{\longrightarrow}\!\varphi_{s}'}$),
 can also be managed (and obtained) by $\widetilde{\EuScript{M}}$.
For $\widetilde{\EuScript{M}}$ to find the \textgreek{l}' part of a ``contextual relation'' 
with \textsigma~given as a high-level idea, it needs to abstract the high-level context
Thus, assuming that $\widetilde{\EuScript{M}}$ has a computable abstraction 
of \textgreek{l}, it needs only to find a computable abstraction of the high-level 
context \textsigma~, and then find the the linguistic construction(s) that can 
be related to both abstractions (i.e., the concrete representation of \textgreek{l} 
and of \textsigma) through any of their related objects. The resulting linguistic 
construction(s) correspond to \textgreek{l}'.

\section{Conclusions and Future Work}
The purpose of this paper has been to establish the basic 
theoretical foundations for the construction of computable 
abstractions and the definition of abstractional machines, which 
aim to provide a solution to the limitations that computers
exhibit under current approaches in performing intellectual 
tasks with results comparable to humans' results.
The detailed study of the applicability of abstractional machines 
is left for future work. 
However, their potential can be observed already.
For instance, if we consider the ability of handling complexity, 
we can notice that by making sense out of an operation, 
an abstractional machine can use it, according to some 
understood purpose (e.g., to build other operations), 
without the need to have or keep available its specification. 
Furthermore, in performing a process, the machine can 
make sense out of the operations that are redundant or unnecessary, 
if any, and skip them, when possible.
Future work is expected to explore relevant aspects of the
new model presented,
such as how to define an abstractional machine that can simulate 
any abstractional process (i.e., a universal abstractional machine), 
as well as to study the applicability of abstractional machines to 
concrete problems for which finding a purely algorithmic solution 
has proved to be hard or impossible. 

\bibliographystyle{abbrv}
\bibliography{sigproc} 

\end{document}